%% file: ms.tex
\documentclass[11pt]{article}
\pdfoutput=1
\usepackage[letterpaper,margin=1in]{geometry}
\usepackage[auth-sc,affil-sl]{authblk}
\usepackage{times}
\usepackage{graphicx}
\usepackage{subcaption}
\usepackage{amsmath, amssymb, amsfonts,bm}
\usepackage{amsthm}

\usepackage{booktabs}
\usepackage{adjustbox}
\usepackage{xcolor}

\usepackage{hyperref}
\hypersetup{
	colorlinks,
	breaklinks,
	linkcolor={red!50!black},
	citecolor={blue!50!black},
	urlcolor={blue!80!black},
}

\usepackage[hyperpageref]{backref}
\usepackage{cleveref}
\crefname{section}{Section}{Sections}
\crefname{figure}{Figure}{Figures}
\crefname{table}{Table}{Tables}
\crefname{appendix}{Appendix}{Appendices}
\crefname{definition}{Definition}{Definitions}

\newtheorem*{definition}{Definition}

\usepackage{enumitem}
\setlist{leftmargin=*} 

\newcommand{\Reals}{\mathbb{R}}
\newcommand{\cD}{\mathcal{D}}
\newcommand{\cT}{\mathcal{T}}
\newcommand{\cX}{\mathcal{X}}
\newcommand{\cY}{\mathcal{Y}}
\newcommand{\cQ}{\mathcal{Q}}
\newcommand{\cV}{\mathcal{V}}
\newcommand{\cF}{\mathcal{F}}

\renewcommand{\vec}[1]{\mathbf{#1}}

\newcommand{\imatch}{g^{\mathrm{v}}}
\newcommand{\mmatch}{g^{\mathrm{m}}}

\newcommand{\doadd}{h^{\mathrm{a}}}
\newcommand{\doreplace}{h^{\mathrm{r}}}

\newcommand{\rhead}{\vec{rh}}
\newcommand{\whead}{\vec{wh}}

\newcommand{\uX}{\underline{X}}

\begin{document}

\title{Transfer Learning in Visual and Relational Reasoning}

\renewcommand\Authands{ and }
\author[1]{T.S. Jayram}
\author[2]{Vincent Marois}
\author[3]{Tomasz Kornuta}
\author[ ]{Vincent Albouy\thanks{Work done while at IBM Research AI, Almaden Research Center}}
\author[*]{\\Emre Sevgen}
\author[1]{Ahmet Ozcan}
\affil[1]{IBM Research AI, Almaden Research Center, San Jose, CA, USA.}
\affil[2]{IBM Research AI, Thomas J. Watson Research Center, Yorktown, NY, USA.}
\affil[3]{NVIDIA, Santa Clara, CA, USA}

\date{}

\maketitle
\newcommand\blfootnote[1]{%
	\begingroup
	\renewcommand\thefootnote{}\footnote{#1}%
	\addtocounter{footnote}{-1}%
	\endgroup
}

\blfootnote{E-mail: {\scriptsize \textit{\{jayram,asozcan\}@us.ibm.com,  \{vincent.marois,vincent.albouy\}@ibm.com, tkornuta@nvidia.com, sesevgen@gmail.com}}}

\begin{abstract}
	Transfer learning has become the de facto standard in computer vision and natural language processing, especially where labeled data is scarce. Accuracy can be significantly improved by using pre-trained models and subsequent fine-tuning. In visual reasoning tasks, such as image question answering, transfer learning is more complex. In addition to transferring the capability to recognize visual features, we also expect to transfer the system's ability to reason. Moreover, for video data, temporal reasoning adds another dimension. In this work, we formalize these unique aspects of transfer learning and propose a theoretical framework for visual reasoning, exemplified by the well-established CLEVR and COG datasets. Furthermore, we introduce a new, end-to-end differentiable recurrent model (SAMNet), which shows state-of-the-art accuracy and better performance in transfer learning on both datasets. The improved performance of SAMNet stems from its capability to decouple the abstract multi-step reasoning from the length of the sequence and its selective attention enabling to store only the question-relevant objects in the external memory.
\end{abstract}

\input{introduction}

\input{related_work}

\input{samnet}

\input{transfer_learning}

\input{experiments}
\input{experiments_reasoning_transfer}

\input{summary}

\bibliographystyle{alpha}
\bibliography{bibliography}

\newpage
\appendix

\input{appendix}

\end{document}

%% file: introduction.tex
\section{Introduction}
In recent years, neural networks, being at the epicenter of the Deep Learning~\cite{lecun2015deep} revolution, became the dominant solutions across many domains, from Speech Recognition~\cite{graves2013speech}, Image Classification~\cite{krizhevsky2012imagenet}, Object Detection~\cite{redmon2016you}, to Question Answering~\cite{weston2014memory} and Machine Translation~\cite{bahdanau2014neural}, among others.
Neural networks, being statistical models~\cite{ripley1993statistical,warner1996understanding}, rely on the assumption that training and testing samples are independent and identically distributed (\textit{iid}).
However, in many real-world scenarios, this assumption does not hold. Moreover, as modern neural models often have millions of trainable parameters, training them requires vast amounts of data, which for some domains (e.g., medical) can be very expensive and/or extremely difficult to collect.
One of the widely used solutions is Transfer Learning~\cite{pan2009survey,weiss2016survey}, a technique which enhances model performance by transferring \emph{information} from one domain to another.

In this work we focus on transfer learning in multi-modal reasoning tasks combining vision and language~\cite{mogadala2019trends}. Recently introduced visual reasoning datasets, such as ~\cite{johnson2017clevr,yang2018dataset,song2018explore}, attempt to probe the generalization capabilities of models under various scenarios, trying to quantify the visual complexity of the scene or the compositional complexity of the question.
While these datasets seem to be well suited for transfer learning, they do not tackle that issue directly.
In particular, \cite{johnson2017clevr} introduces the CLEVR dataset along with two variants (CoGenT-A and -B) with varying combinations of visual attributes.
Similarly, COG \cite{yang2018dataset}, a Video QA reasoning dataset, contains two variants (Canonical and Hard) with different number of frames and scene complexity.
Both datasets come with a number of task classes that require similar reasoning abilities (e.g. compare or count objects).
The open question we address in this paper is whether we can build a neural architecture that fosters transfer learning and generalizes from simpler to more complex reasoning tasks.

\noindent The contributions of this paper are the following:
\begin{enumerate}
	\item We introduce a new model called SAMNet (Selective Attention Memory Network).
	The unique features of SAMNet include the dynamic processing of video input frame-by-frame, multi-step reasoning over a given image and remembering the relevant concepts.
	\item We propose a taxonomy of transfer learning to highlight new research directions, identify missing areas and encourage new datasets for transfer learning in visual reasoning. We evaluate the CLEVR/CoGenT and COG tasks to show how they map onto the proposed theoretical framework.
	\item We perform an extensive set of experiments, comparing our model with baselines on the existing tasks, achieving state-of-the-art results on COG~\cite{yang2018dataset}. We also designed two new tasks for reasoning transfer and show that the design of SAMNet promotes transfer learning capabilities and generalization.

\end{enumerate}

%% file: related_work.tex
\section{Related work}
\label{sec:related_work}

In Computer Vision, it is now standard practice to pretrain an image encoder (such as VGG~\cite{simonyan2014very} or ResNet~\cite{he2016deep}) on large-scale datasets (such as ImageNet~\cite{deng2009imagenet}), and reuse the weights in unrelated domains and tasks, such as segmentation of cars~\cite{iglovikov2018ternausnet} or Visual Question Answering (VQA) in a medical domain~\cite{kornuta2019leveraging}.
Such performance improvements are appealing, especially in cases where both the domain (natural vs. medical images) and the task (image classification vs. image segmentation vs VQA) change significantly.

Similar developments have emerged in the Natural Language Processing (NLP) community.
Using shallow word embeddings, such as word2vec~\cite{mikolov2013distributed} or GloVe~\cite{pennington2014glove}, pretrained on large corpuses from e.g.\ Wikipedia or Twitter, has been a standard procedure across different NLP domains and tasks.
Recently, there is a clear, growing trend of utilization of deep contextualized word representations such as ELMo~\cite{peters2018deep} (based on bidirectional LSTMs~\cite{hochreiter1997long}) or BERT~\cite{devlin2018bert} (based on the Transformer~\cite{vaswani2017attention} architecture), where entire deep networks (not just the input layer) are pretrained on very large corporas.

Visual Reasoning tasks, combining both the visual and language modalities~\cite{mogadala2019trends}, naturally draw from those findings by reusing the pretrained image and word/question encoders.
Ensuring models can generalize on high-level, abstract \emph{reasoning} has been a recent interest of the community.
The first attempt at establishing a framework enabling transfer learning of reasoning skills was CLEVR~\cite{johnson2017clevr} with its two CoGenT variants.
This fostered research~\cite{mascharka2018transparency, perez2018film, johnson2017inferring,marois2018transfer}, proposing models designed to transfer from one domain to another, with different distributions of visual attribute combinations.
Similarly, the baseline model introduced alongside the COG dataset~\cite{yang2018dataset} has shown the ability to solve tasks not explicitly trained on, by leveraging knowledge learned on other tasks.

These results clearly indicate the usefulness of transfer learning, despite the assumption of similar distributions between the source and target domains not being respected.
Conjointly, transfer learning raises several research questions, such as the characteristics which make a whole dataset or a particular task more favorable to be used in pretraining (notably ImageNet~\cite{huh2016makes}), or regarding the observed performance correlation of models with different architectures between the source and target domains~\cite{kornblith2019better}.
One of the most systematic works in this area is the computational taxonomic map for task transfer learning~\cite{zamir2018taskonomy}, aiming at discovering the dependencies between twenty-six different computer vision tasks.
In this work, we extend this idea further and introduce a taxonomy allowing us to isolate and quantify different aspects of visual reasoning.



%
%

%% file: samnet.tex
\section{Selective Attention Memory (SAM) Network}

\begin{figure*}
	\centering
	\includegraphics[width=\textwidth]{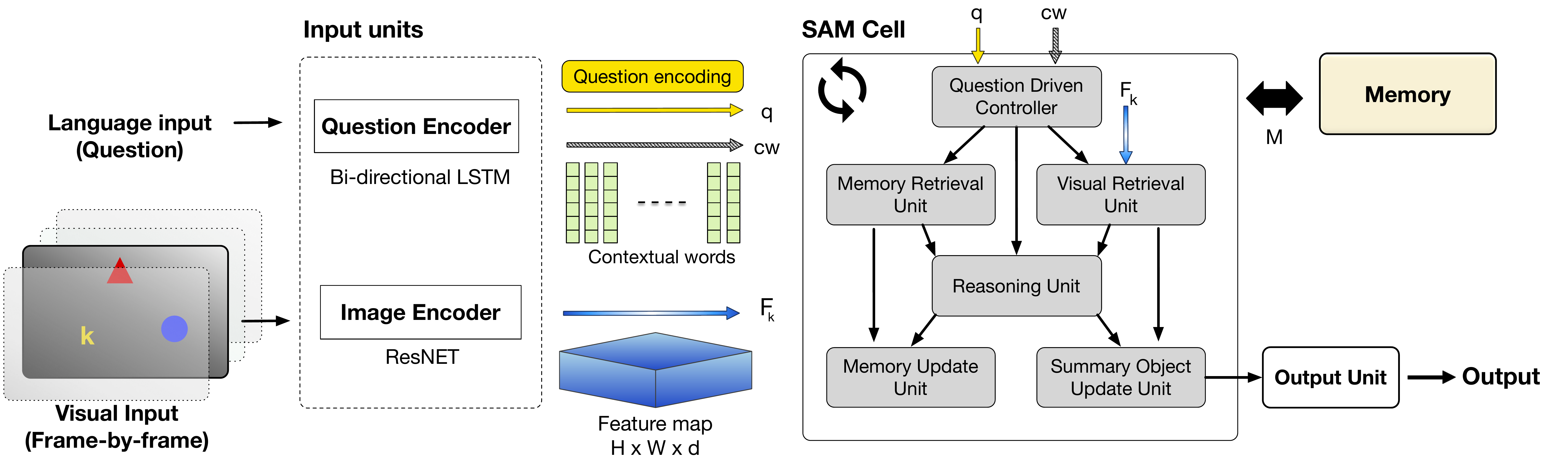}
	\caption{General architecture of SAMNet.}
	\label{fig:samnet}\vspace{-10pt}
\end{figure*}

SAM Network (SAMNet) is an end-to-end differentiable recurrent model equipped with an external memory~(\cref{fig:samnet}). At the conceptual level, SAMNet draws from two core ideas:
iterative reasoning as proposed e.g. in MAC (Memory-Attention-Composition) Network~\cite{hudson2018compositional,marois2018transfer} and use of an external memory, as in Memory-Augmented Neural Networks such as NTM (Neural Turing Machine)~\cite{graves2014neural}, DNC (Differentiable Neural Computer)~\cite{graves2016hybrid} or DWM (Differentiable Working Memory)~\cite{jayram2018learning}.

A distinctive feature of the SAM Network is its frame-by-frame temporal processing approach, where a single frame can be accessed at once. This is a notable difference from~\cite{haurilet2019s}, which uses graph traversal reasoning. The recurrent nature of SAMNet does not prevent frame sequences longer than those used for training.
The memory locations store relevant objects representing contextual information about words in text and visual objects extracted from video.
Each location of the memory stores a $d$-dimensional vector. 
The memory can be accessed through either content-based addressing, via dot-product attention, or location-based addressing. Using gating mechanisms, correct objects can be retrieved in order to perform multi-step spatio-temporal reasoning over text and video.
A notable feature of this design is that the number of addresses $N$ can be changed between training and testing, to fit the data characteristics.


\smallskip

\noindent\textbf{Input Units.}
The input to SAMNet consists of a question component and a visual component that is either a single frame in case of a static image or a sequence of frames in case of a video.
The question is represented as a sequence of word embeddings (either one-hot or preprocessed by GloVe~\cite{pennington2014glove}) corresponding to words in the question.
Each frame of the visual input is encoded using raw pixels or preprocessed by a pre-trained deep network such as RESNet~\cite{he2016deep}.
The language input of $L$ word embeddings is processed by a Bi-LSTM based question encoder that produces \emph{contextual words} $\vec{cw} \in \Reals^{L \times d}$.
The final states in either direction of the BI-LSTM are also concatenated and passed through a linear layer to produce a global representation of the question 
$\vec{q} \in \Reals^d$, called the \emph{question embedding}.
The $k$-th frame of the visual input, where $k$ indexes over the frames in temporal order, is processed by an image encoder---a CNN with appropriately chosen number of layers---to produce a \emph{feature map} 
$\vec{F}_k \in \Reals^{H \times W \times d}$.

\smallskip

\noindent\textbf{The SAM Cell.} The contextual words, question embedding and frame-by-frame feature map form the inputs to SAMNet's core, namely, a recurrent cell called the SAM Cell. Unrolling a new series of $T$ cells for each frame allows $T$ steps of iterative reasoning, similar to~\cite{hudson2018compositional}. Information flows between frames through the external memory.
During the $t$-th reasoning step, for $t=1,2, \dots, T$, SAM Cell maintains the following information as part of its recurrent state:
(a) $\vec{c}_t \in \Reals^d$, the control state used to drive reasoning over objects in the frame and memory; and
(b) $\vec{so}_t  \in \Reals^d$, the summary visual object representing the relevant object for step $t$.
Let $\vec{M}_t \in  \Reals^{N \times d}$ denote the external memory with $N$ slots at the end of step $t$.
Let $\whead_t \in  \Reals^N$ denote an attention vector over the memory locations;
in a trained model, $\whead_t$ points to the location of the first empty slot in memory for adding new objects.

\smallskip

\noindent\textbf{Question-driven Controller.}
This module drives attention over the question to produce $k$ control states, one per reasoning operation.
The control state $\vec{c}_t$ at step $t$ is then fed to a \emph{temporal classifier},
a two-layer feedforward network with ELU activation in the hidden layer of $d$ units.
The output $\bm{\tau}_t$ of the classifier is intended to represent the different temporal contexts (or lack thereof) associated with the word in focus for that reasoning step.
For the COG dataset, we pick 4 classes to capture the terms labeled ``last'', ``latest'', ``now'', and ``none''.

\smallskip

\noindent\textbf{Retrieval Units.} The visual retrieval unit uses the information generated above to extract a relevant object $\vec{vo}_t$ from the frame.
A similar operation on memory yields the object $\vec{mo}_t$. The memory operation is based on attention mechanism, and resembles content-based addressing. Therefore, we obtain an attention vector over memory addresses that we interpret to be the \emph{read head}, denoted by $\rhead_t$.

\smallskip

\noindent\textbf{Reasoning Unit.}
This module is the backbone of SAMNet, which determines the gating operations to be performed on the external memory, as well as determining the correct object's location for reasoning.
To resolve whether we have a valid object from the frame (and similarly for memory), we execute the following reasoning procedure.
First, we compute a simple aggregate\footnote{%
	This is closely related to R\'{e}nyi entropy and Tsallis entropy of order~2.} of the visual attention vector $\vec{va}_t$ of dimension $L$ ($L$ denotes the number of feature vectors for the frame):
$vs_t = \sum_{i=1}^L [\vec{va}_t(i)]^2$. It can be shown that the more localized the attention
vector, the higher the aggregate value.
We perform a similar computation on the read head $\rhead_t$ over memory locations.
We input these two values, along with the temporal class weights $\bm{\tau}_t$, into a 3-layer feedforward classifier with hidden ELU units to extract 4 gating values (in $[0,1]$) for the current reasoning step:
(a) $\imatch_t$, which determines whether there is a valid visual object;
(b) $\mmatch_t$, which determines whether there is a valid memory object.
(c) $\doreplace_t$, which determines whether the memory should be updated by replacing a previously stored object with a new one; and
(d) $\doadd_t$, which determines whether a new object should be added to memory.
We stress that the network has to learn via training how to correctly implement these behaviors.

\smallskip

\noindent\textbf{Memory Update Unit.}
The unit first determines the memory location where an object could be added:
\[ \vec{w}_t = \doreplace_t \cdot \rhead_t + \doadd_t \cdot \whead_{t-1} \]
Above, $\vec{w}_t$ denotes the pseudo-attention vector that represents the ``location'' where the memory update should happen.
$\vec{w}_t$ sums up to at most 1 and can be zero, indicating in this case there is no need adding a new object nor replacing an existing object.
We then update the memory accordingly:
\[ \vec{M}_t = \vec{M}_{t-1} \odot (\vec{J} - \vec{w}_t  \otimes \mathbf{1}) + \vec{w}_t  \otimes \vec{vo}_t,\]
where $\vec{vo}_t$ denotes the object returned by the visual retrieval unit.
Here $\vec{J}$ denotes the all ones matrix, $\odot$ denotes the Hadamard product and $\otimes$ denotes the Kronecker product.
Note that the memory is unchanged in the case where $\vec{w}_t = 0$, i.e., $\vec{M}_t = \vec{M}_{t-1}$.
We finally update the write head so that it points to the succeeding address if an object was added to memory or otherwise stay the same.
Let $\whead'_{t-1}$ denote the circular shift to the right of $\whead_{t-1}$ which corresponds to the soft version of the head update.
Then:
\[ \whead_t = \doadd_t \cdot \whead'_{t-1} + (1-\doadd_t) \cdot \whead_{t-1} \]

\begin{figure*}[htbp]
	\centering
	\includegraphics[width=0.9\textwidth]{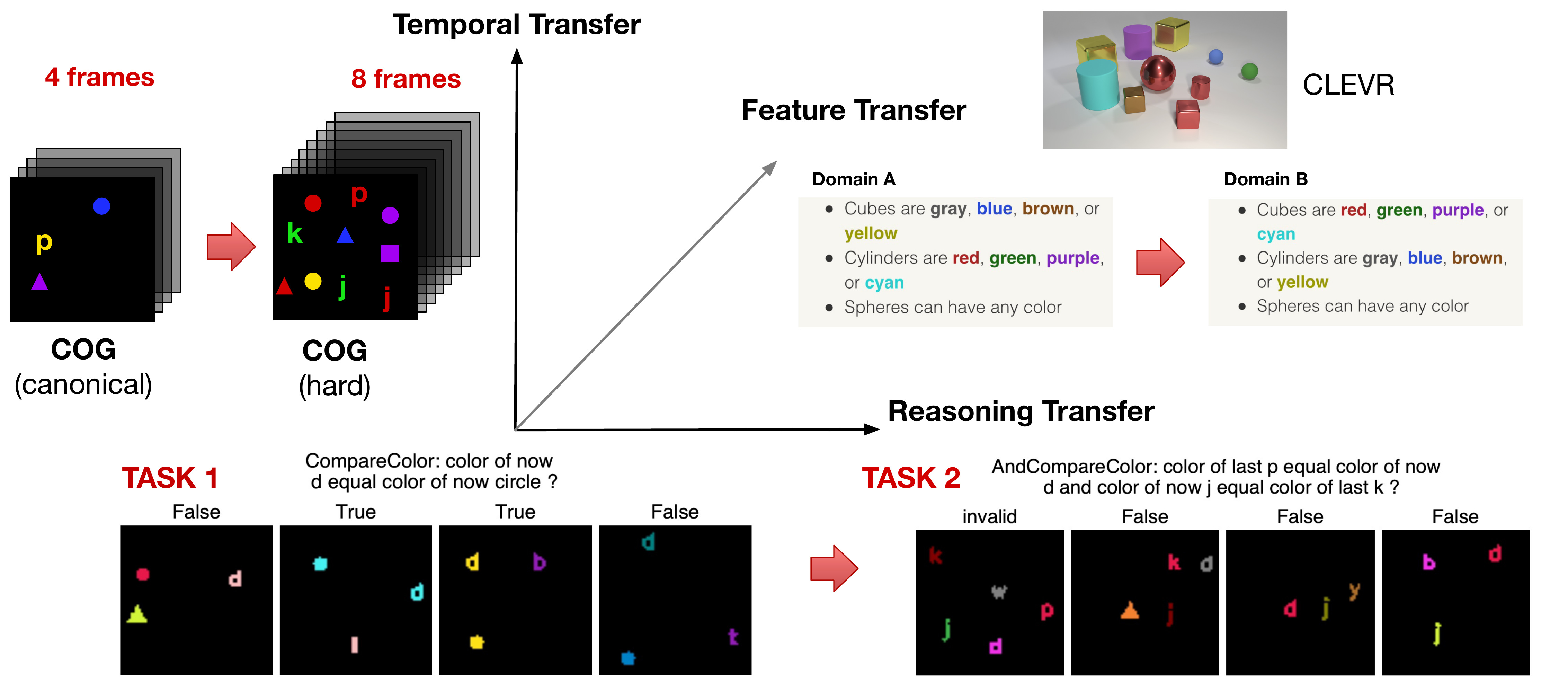}
	\caption{Transfer learning taxonomy.}
	\label{fig:taskonomy}\vspace{-10pt}
\end{figure*}

\noindent\textbf{Summary Update Unit.}
This unit updates the (recurrent) summary object to equal the outcome of the $t$-th reasoning step.
We first determine whether the relevant object $\vec{ro}_t$ should be obtained from memory or the frame according to:
\[ \vec{ro}_t = \imatch_t \cdot \vec{vo}_t + \mmatch_t \cdot \vec{mo}_t \]
Note that $\vec{ro}_t$ is allowed to be a null object (i.e. 0 vector) in case neither of the gates evaluate to true.

\smallskip

\noindent\textbf{Output Unit.} Finally, $\vec{so}_t$ is the output of a linear layer whose inputs are $\vec{ro}_t$ and the previous summary object $\vec{so}_{t-1}$.
This serves to retain additional information in $\vec{so}_{t-1}$, e.g., if it held the partial result of a complex query with Boolean connectives.

%% file: transfer_learning.tex
\section{Transfer Learning}
\label{sec:transfer_learning}
Due to the complex nature of visual reasoning, it appears that
transfer learning from a source domain to a target domain can be investigated in
multiple ways, depending on how the two domains are related. This includes visual attributes such as size, color and shape, temporal characteristics reflected in the number of frames and scene complexity, and finally the reasoning difficulty. 
To establish a formal framework for these various aspects, we first recall the basic theoretical notions in transfer learning~\cite{pan2009survey}.

A \emph{domain} is a pair $\cD = (\cX,P(X))$, where $\cX$ is a feature space and $P(X)$ is a marginal probability distribution.
For visual reasoning problems considered in this paper,
$\cX$ will consist of purely visual inputs, i.e., either images or videos in some cases, or
a combination of both visual inputs and questions in other cases.
A \emph{task} is a pair $\cT= (\cY,f(\cdot))$, where $\cY$ is a label space and $f: \cX \to \cY$ is a prediction function.
When the domain elements consist of both the question and the visual input, there is only one task, namely, to answer the
question\footnote{%
	For the COG dataset, the answer is a tuple, one for each frame in the video, whereas for typical video answering datasets,
	only a single answer is needed for the entire video.}. %
If the domain elements consist of just the visual inputs, then the task is defined by the question so that each question
defines a separate task.

\begin{definition}[\cite{pan2009survey}]
	\label{defn:transfer}
	Given a source domain $\cD_S$ and a source learning task $\cT_S$, a target domain $\cD_T$ and a target learning task $\cT_T$, transfer learning aims to help improve the
	learning of the target predictive function $f_T(\cdot)$ in $\cD_T$ using the knowledge  in $\cD_S$ and $\cT_S$, where $\cD_S \ne \cD_T$, or $\cT_S \ne \cT_T$.
\end{definition}
In all our applications, $\cX_S = \cX_T$, so $\cD_S \ne \cD_T$ means that the marginal distributions $P_S$ and $P_T$ are different.
Similarly, $\cT_S \ne \cT_T$ means that either $Y_S \ne Y_T$ or that the associated prediction functions are different.

Although~\cref{defn:transfer} is quite general, it does not adequately capture all artifacts present in visual reasoning.
For example, consider the transfer learning setting where the tasks $\cT_S$ and $\cT_T$ are the same
but the marginal distributions $P_S$ and $P_T$ are different (referred to as \emph{domain adaptation}).
As mentioned in the introduction, one setting is the case of static images,
where this could be due to having different feature combinations in the source and target.
A different setting is in the context of video reasoning where the number of frames can increase significantly going from source to target.
These require possibly very different methods: the first involves building disentangled feature representations that can generalize across
domains; the second might need external memory to remember relevant objects to generalize across frame lengths.
Another situation is when the questions themselves can be grouped into families such as count-based queries,
comparison of objects, or existence of objects with certain features etc.
This entails studying transfer learning between families of tasks which requires extending the above definition.

\begin{table}[b!]
	\centering
	\begin{tabular}{cccc}
		\toprule
		Variant	& Cubes	& Cylinders &	Spheres	\\
		\midrule
		CLEVR &  Any Color  & Any Color 	&	Any color  \\
		CoGenT-A &  Family A  & Family B 	&	Any color  \\
		CoGenT-B	&	Family B  &	Family A	&	Any color \\
		\bottomrule
	\end{tabular}
	\caption{Allowed feature combinations in the CoGenT-A and CoGenT-B variants of the CLEVR dataset.}
	\label{tab:cogent_conditions}
\end{table}

We now formally define 3 kinds of transfer learning problems, namely,
\emph{feature transfer}, \emph{temporal transfer},
and \emph{reasoning transfer}.
These are illustrated in~\cref{fig:taskonomy} using representative examples from CLEVR-CoGenT and COG. These datasets were chosen because they 
are particularly suited for experimental investigations of transfer learning described in the following section.
Let $\cQ$ and $\cV$ denote the set of questions and visual inputs, respectively.
\begin{description}
	\item[Feature Transfer:] In this setting of domain adaptation, $\cX_S = \cX_T \subseteq \cQ \times \cV$
	and the task $f(q,v)$ is just the answer to the question $q$ on visual input $v$. The output set $\cY$ is the union of legitimate answers
	over all questions in $\cQ$.
	The marginal distributions $P_S$ and $P_T$ differ in the feature attributes such as shape, color, and size, or their combinations
	thereof.

	\item[Temporal Transfer:] This setting is similar to attribute adaptation in that $\cX_S = \cX_T \subseteq \cQ \times \cV$
	and there is a single task.
	The key difference is that we introduce a notion of complexity $C(v) = (n, m)$ for a visual input $v$,
	where $n$ equals the maximum number of objects $n$ in an image, and $m$
	equals  the number of frames in a video.
	For any visual input $v_S$ coming from $\cX_S$ with $C(v_S) = (n_S, m_S)$
	and for any visual input $v_T$ coming from $\cX_T$ with $C(v_T) = (n_T, m_T)$, we require that $n_T \ge n_S$ and
	$m_T \ge m_S$ with at least one inequality being a strict one.
	Thus, we necessarily increase the complexity of the visual input going from the source to the target domain.

	\item[Reasoning Transfer:]
	This setting requires an extension of~\cref{defn:transfer} above to investigate transfer learning when
	grouping questions into families. Let $\cV$ be the feature space consisting of visual inputs only, shared by
	all tasks, with a common marginal distribution $P(X)$. For each question $q \in \cQ$, we define the task
	$\cT_q = (\cY_q, f_q(\cdot))$ where
	the output set $\cY_q$ is the set of legitimate answers to $q$ and $f_q(v)$, for a visual input $v$,
	is the answer to question $q$ on visual input $v$.
	Thus, tasks are in a 1-1 correspondence with questions.
	A \emph{task family} is a probability distribution on tasks which in our case can be obtained by defining the distribution on $\cQ$.
	Given a task family, the goal is to learn a prediction function that gives an answer to $f_q(v)$ for $v \in \cV$ chosen according
	to the feature space distribution and $q$ chosen according to the task probability distribution.
	Suppose $\cF_S$ is the source task family and $\cF_T$ is the target task family.
	Transfer learning aims to help improve the learning of the predictive function for the target task family
	using the knowledge in the source task family.

\end{description}

If labeled data is available for $\cX_T$, a training algorithm distinction we make is between \emph{zero-shot learning} and \emph{finetuning}. Finetuning entails the use of labeled data in the target domain $\cD_T$, foreseeing performance gain on the target task $\cX_T$, after initial training on $\cX_S$ and additional training on $\cX_T$. Zero-shot learning thus refers to immediate test on $\cX_T$ after initial training on $\cX_S$.

%% file: experiments.tex
\section{Experiments}
\label{sec:experiments}

We implemented and trained our SAMNet model using MI-Prometheus~\cite{kornuta2018accelerating}, a framework based on PyTorch~\cite{paszke2017automatic}.
We evaluated the model on
the CLEVR dataset~\cite{johnson2017clevr}, a diagnostic dataset for Image Question Answering and on the COG dataset~\cite{yang2018dataset}, a video reasoning~\cite{mogadala2019trends} dataset developed for the purpose of research on relational and temporal reasoning.
We briefly describe both datasets in the following two subsections.
In all experiments, we trained SAMNet using $k = 8$ reasoning steps and external memory with $N = 8$ slots, each storing an array of 128 floats.
When working with CLEVR, we deactivated SAMNet's external memory and temporal-related modules as they are unnecessary while reasoning about static images. 
Additional information regarding the dataset parameters 
can be found in~\cref{sec:datasets-desc}.

We start with an evaluation of SAMNet's performance, comparing it to baselines on both CLEVR and COG datasets, without transfer learning.
The rest of the experiments follows the proposed taxonomy.
We first investigate feature transfer using CLEVR-CoGenT, followed by temporal transfer on the COG variants. Finally, we assess the reasoning transfer capability of SAMNet on both datasets, by reorganizing groups of questions into new training and test splits.

\subsection{Performance of SAMNet on CLEVR}
\label{sec:clevr-baseline-compare}
CLEVR images contain objects characterized by a set of attributes (shape, color, size and material). The questions are grouped into 5 categories: \textit{Exist}, \textit{Count}, \textit{CompareInteger}, \textit{CompareAttribute}, \textit{QueryAttribute}.
The CLEVR authors also introduced CLEVR-CoGenT (Constrained Generalization Test), containing 2 variants, CoGenT-A and CoGenT-B, with varying combinations of color-shape attributes.
The colors are partitioned into two complementary families:
Gray, Blue, Brown and Yellow in Family A; and Red, Green, Purple, Cyan in Family B.
The cubes and cylinders take colors from complementary families in each variant with opposite configurations; the spheres can take any color in both variants (See~\cref{tab:cogent_conditions}).
As the input domain consist of the set of objects with all attribute values, both variants differ by their marginal distributions $P_S$ and $P_T$.

\begin{figure}[hbtp]
	\centering
	\includegraphics[width=0.5\textwidth]{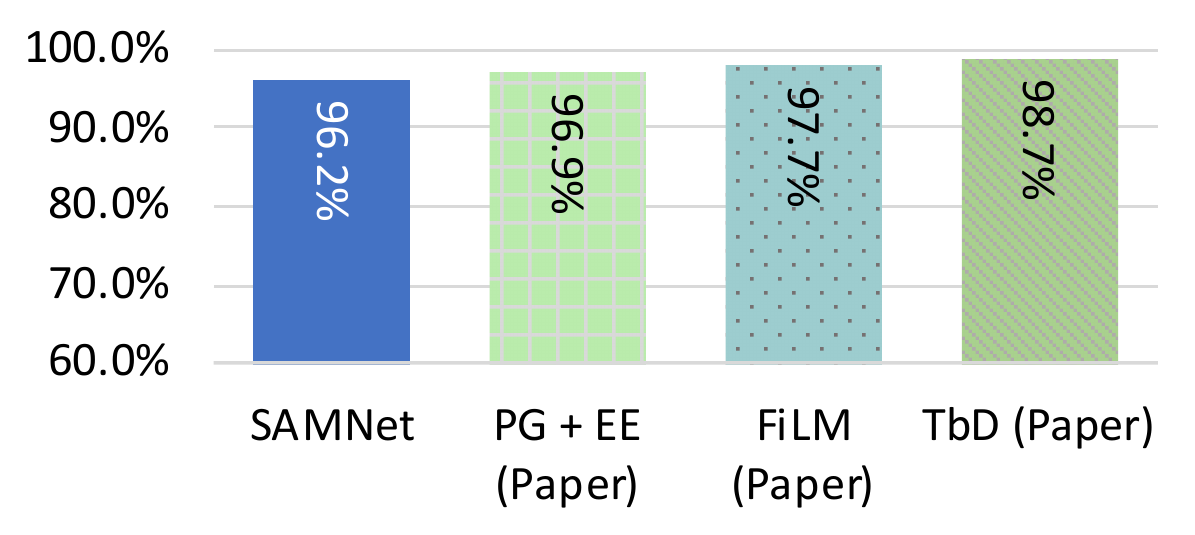}
	\caption{Comparison of SAMNet to baselines on CLEVR.}
	\label{fig:clevr_baselines}
\end{figure}

\Cref{fig:clevr_baselines} presents the accuracy achieved by SAMNet when compared to selected state-of-the-art models: Transparency by Design networks (TbD)~\cite{mascharka2018transparency}, Feature-wise Linear Modulation (FiLM)~\cite{perez2018film} and the Program Generator + Execution Engine (PG + EE)~\cite{johnson2017inferring}.
SAMNet reaches a comparable 96.2\% accuracy, with TbD achieving the best score of 98.7\%.

\subsection{Performance of SAMNet on COG}
\label{sec:cog-baseline-compare}

COG dataset contains short videos, with associated questions grouped into 23 categories\footnote{COG also proposes a pointing version of the tasks, which are not considered in this work.}.
COG comes in two variants: Canonical (easy) and Hard, differing mainly on the number of frames in the video, the maximum amount of look-back in frame history containing relevant information for reasoning, and the number of object distractors present in a given frame (see~\cref{tab:cog_variants}).

\begin{table}[htbp]
	\centering
	\begin{tabular}{lccc}
		\toprule
		Variant	& Frames & History	& Distractors \\
		\midrule
		Canonical (Easy) & 4 & 3 & 1\\
		Hard  & 8 & 7 & 10\\
		\bottomrule
	\end{tabular}
	\caption{Details of the Canonical and Hard variants of COG.}
	\label{tab:cog_variants}
\end{table}

\begin{figure}[hbtp]
	\centering
	\includegraphics[width=0.7\textwidth]{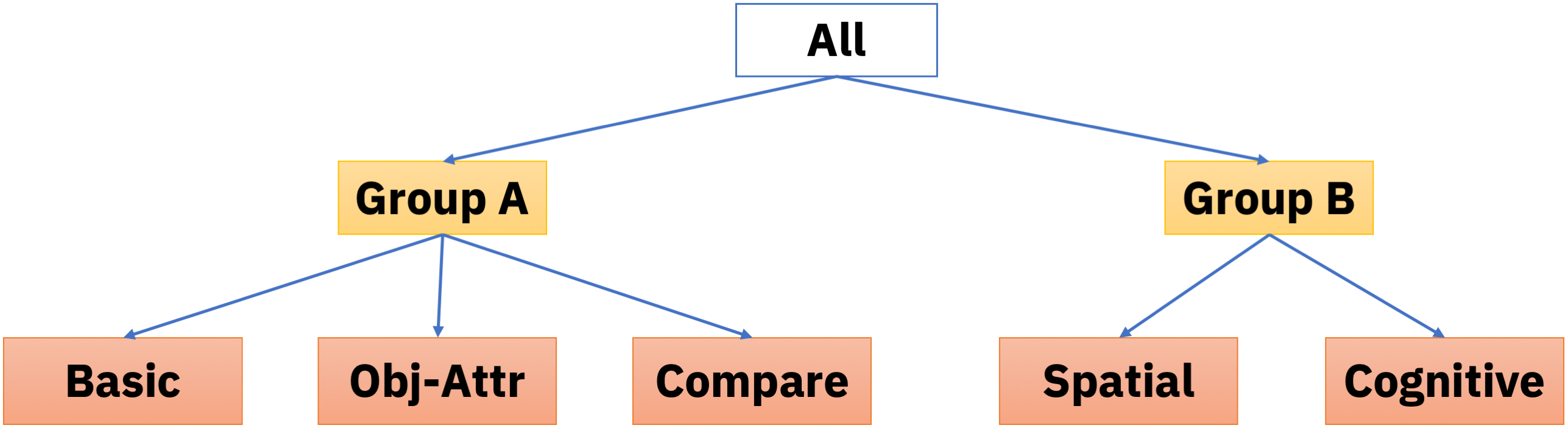}
	\caption{Hierarchy of Task Groups in the COG dataset.}
	\label{fig:task-groups}
\end{figure}

Since the number of task classes is large, we designed a 2-level hierarchy of task groups using the
description of tasks, as shown in~\cref{fig:task-groups}.

For groups at the lowest level, we chose the following task classes to be placed in those groups.
Below, substitute each of \textit{Shape} and \textit{Color} for  \uX{} to obtain the task class.
\begin{description}
	\item[Basic:] \textit{Exist}\uX, \textit{Get}\uX{} and \textit{Exist};
	\item[Obj-Attr:] \emph{SimpleCompare}\uX{} and \textit{AndSimpleCompare}\uX;
	\item[Compare:] \textit{Compare}\uX,  \textit{AndCompare}\uX{} \& \textit{Exist}\uX\textit{Of};
	\item[Spatial:] \textit{ExistSpace}, \textit{Exist}\uX\textit{Space}, and \textit{Get}\uX\textit{Space};
	\item[Cognitive:] \textit{ExistLastColorSameShape}, \textit{ExistLastShapeSameColor} and \textit{ExistLastObjectSameObject}
\end{description}

\begin{figure}[htbp]
	\centering
	\begin{subfigure}{\textwidth}
		\centering
		\includegraphics[width=\textwidth]{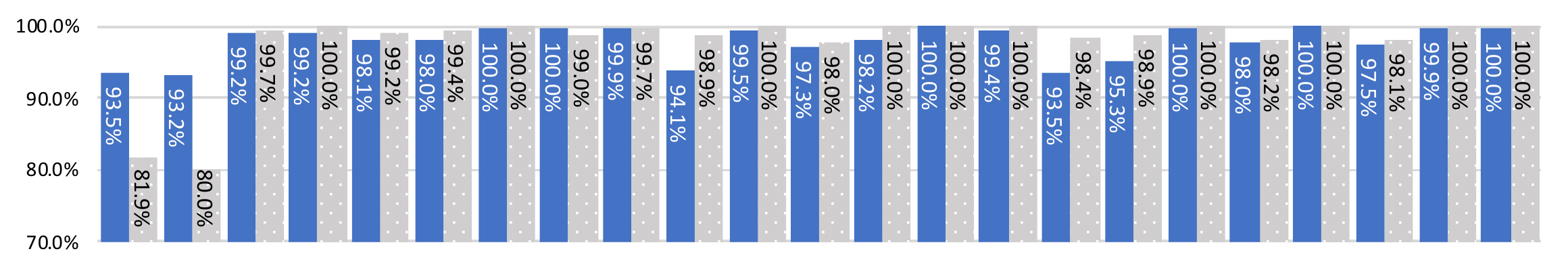}
	\end{subfigure}%
	\newline
	\begin{subfigure}{\textwidth}
		\centering
		\includegraphics[width=\textwidth]{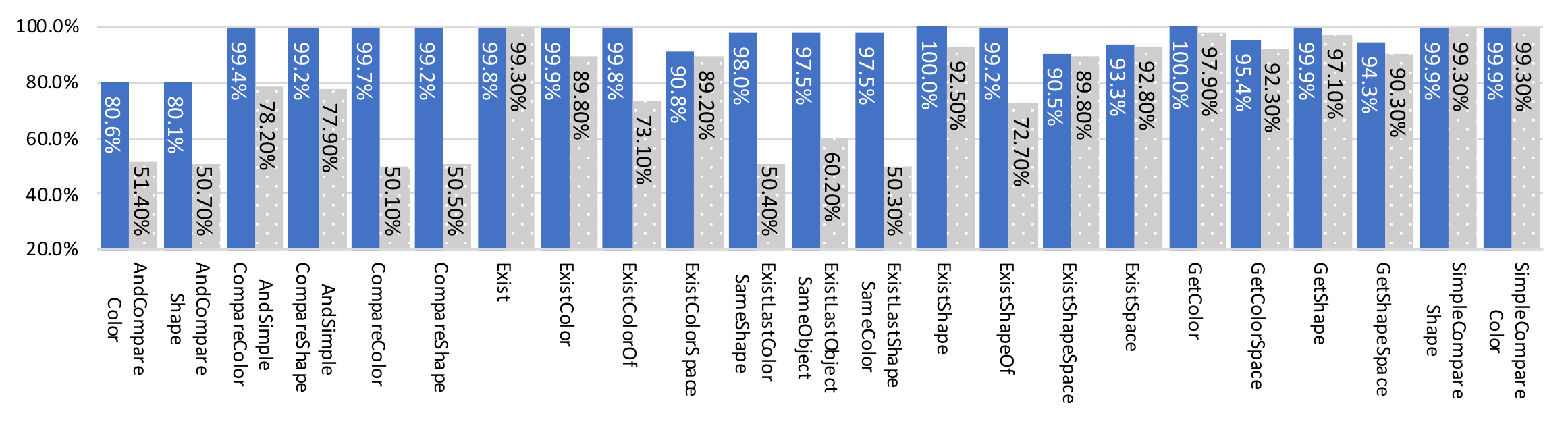}
	\end{subfigure}%
	\caption{Comparison of test set accuracies of SAMNet (blue) with original results achieved by the baseline model~\cite{yang2018dataset} (dotted gray) on Canonical (top) and Hard (bottom) variants of the COG dataset.}
	\label{fig:samnet_cog_detailed}\vspace{-10pt}
\end{figure}

We compared our results with the baseline model introduced in~\cite{yang2018dataset}, the same paper as the COG dataset.
The most important results are highlighted in~\cref{fig:samnet_cog_detailed}; full comparison can be found in~\cref{sec:cog-all-results}.

For the Canonical variant (top row), we achieve similar accuracies for the majority of tasks (with a total average accuracy of 98.0\%, compared to 97.6\% for the baseline model), with significant improvements (around 13 points) for \textit{AndCompare} tasks.
As these tasks focus on compositional questions referring to two objects, we hypothesize that our model achieves better accuracy due to its ability to selectively pick and store relevant objects from the past frames in its external memory.
For the Hard variant, we achieve a total average accuracy of 96.1\% compared to 80.1\% for the baseline model, demonstrating that our model can adapt to larger number of frames and distractors.
SAMNet improves upon the baseline model on all tasks, with improvements varying from 0.5 to more than 30 points, especially outperforming in the most complex tasks (\textit{AndCompare}\uX, \textit{Compare}\uX).

\subsection{Feature transfer using CLEVR-CoGenT}
\label{sec:feature}

\begin{figure}[htbp]
	\centering
	\includegraphics[width=0.9\textwidth]{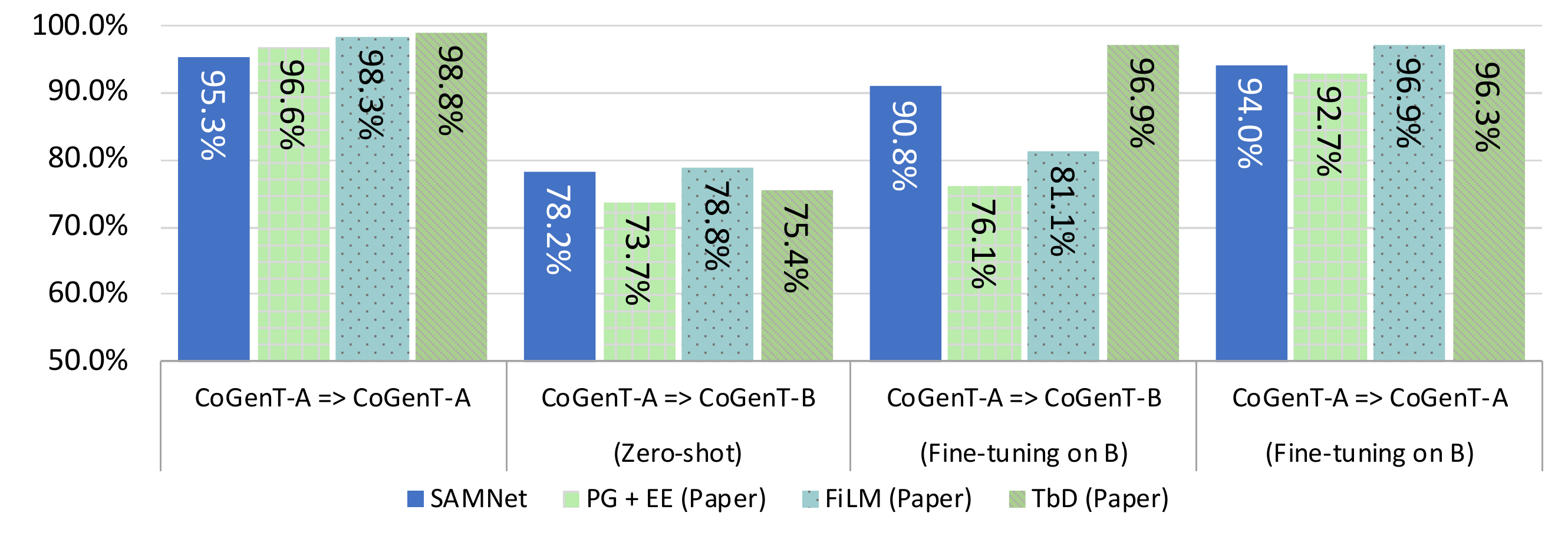}
	\caption{Feature transfer: test accuracies when transferring between CoGenT-A and CoGenT-B.}
	\label{fig:CoGenT-B-results}
\end{figure}

In order to quantify SAMNet's feature transfer ability, we used the CoGenT variants of CLEVR and performed a set of experiments, starting from training on CoGenT-A, followed by:
an immediate test (zero-shot learning) on CoGenT-B; and fine-tuning for a single epoch on 30k samples from CoGenT-B before testing (following the methodology used in~\cite{johnson2017inferring, mascharka2018transparency, perez2018film, marois2018transfer}).
In \cref{fig:CoGenT-B-results}, we compare the achieved performance with the previously selected state-of-the-art models.
We also consider a direct test on CoGenT-A; and a test on CoGenT-A after fine-tuning on CoGenT-B.

When training and testing on CoGenT-A, SAMNet performs slightly worse than the other models, matching the results observed in \cref{sec:clevr-baseline-compare}.
When zero-shot testing on CoGenT-B, all models show relatively good performance (around 75\%), which represents nonetheless a drop of 17 points with respect to CoGenT-A.
All models perform better when fine-tuned on CoGenT-B.
However, this process also results in a degradation of performance on CoGenT-A, with SAMNet and TbD showing an ability to limit this drop in accuracy.
While this suggests both models were able to adapt representations for both domains, these representations are not fully disentangled.

\subsection{Temporal transfer in COG}
\label{sec:temporal}

\begin{figure}[b!]
	\centering
	\includegraphics[width=0.8\textwidth]{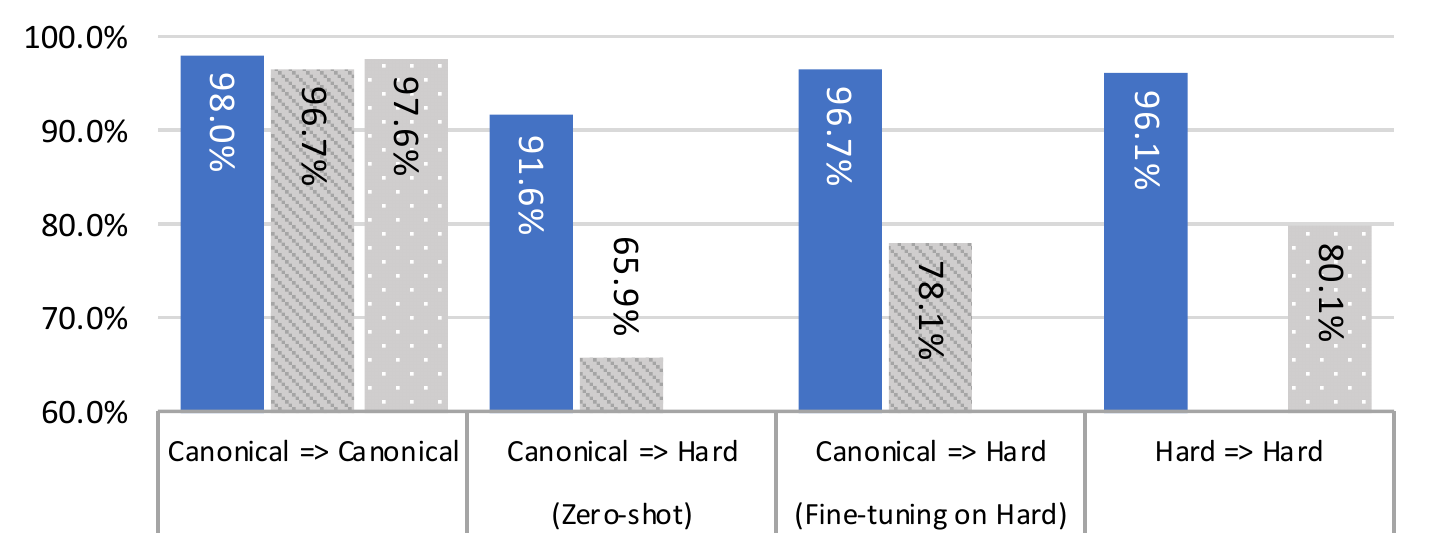}
	\caption{Temporal transfer capabilities of SAMNet (blue) and baseline models (gray -striped and -dotted) when moving from Canonical to Hard  in COG.}
	\label{fig:samnet_cog_overall_transfer}
\end{figure}

Temporal transfer tests the transfer learning ability regarding the frame sequence length, frame history required for reasoning, and the number of distracting objects.
Therefore, we train the models on the Canonical variant of COG dataset and test them on the Hard variant~(\cref{fig:samnet_cog_overall_transfer}).
As the original paper~\cite{yang2018dataset} didn't provide such experiments, we supplement the original accuracies (denoted as gray-dotted) with the ones achieved by the COG baseline model that we have obtained by running the original code (gray-striped) provided by the authors~\cite{yang2018implement}.

The first column displays the aggregated scores when training and testing on the Canonical variant.
The close scores of both base models (gray striped and dotted) highlight the faithful reproduction of the baseline model.
For both cases of zero-shot learning and fine-tuning using a single epoch, SAMNET significantly outperformed the baseline model, by 25 and 17 points respectively.
This might partially results from the frame-by-frame processing of videos executed by SAMNet by design, as opposed to the processing of aggregated frames performed by the baseline.
Interestingly, fine-tuning yields a mild boost of +0.6\% on the accuracy achieved by the model trained exclusively on the Hard variant.
This suggests that SAMNet managed to develop elementary skills (such as memory usage and attention on relevant entities) and successfully uses them when dealing with longer videos and more complex scenes.


%% file: experiments_reasoning_transfer.tex
\subsection{Reasoning transfer in CLEVR-CoGenT}
\label{sec:reasoning-transfer-clevr}
In the following experiments, we used CoGent-A only.
Using the question groups defined by the authors, we organized training and testing splits with the goal of measuring whether mastering reasoning for one questions group can help learning others.

\begin{figure}[htbp]
	\centering
	\includegraphics[width=0.8\textwidth]{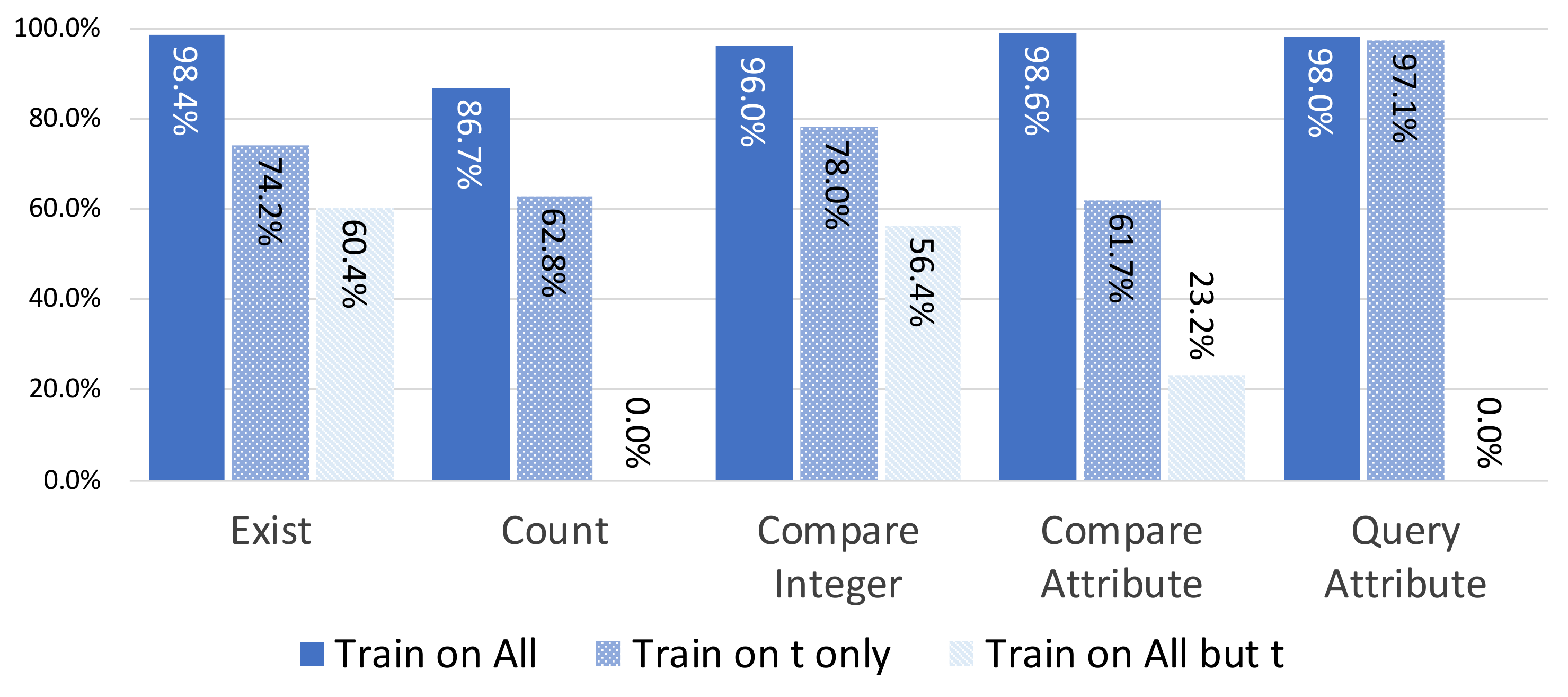}
	\caption{Accuracies of SAMNet when testing its reasoning transfer capabilities on the CoGenT-A variant.}
	\label{fig:cogent_reasoning_transfer}
\end{figure}

We first trained a single SAMNet instance on CoGenT-A and measured its accuracy on each of the task group $t$ separately.
Those results, indicated in \cref{fig:cogent_reasoning_transfer} as ``Train on All'', serve as baselines for the next two sets of experiments.

We next trained and tested 5 SAMNet instances, each on a single task group $t$ (``Train on t only'').
We observe significant accuracy drops (from 18 up to almost 35 points) in all task groups, aside of \textit{Query Attribute}.
We hypothesize \textit{Query Attribute} requires mastering visual grounding and recognizing object attributes, as opposed to the other groups which may not enforce it.
Thus, training on \textit{Query Attribute} helps others, whereas without any training on that particular group, SAMNet battles developing such skill.

Finally, we trained 5 SAMNet instances as follows: for each task group $t$ we trained a single instance on all tasks but $t$, and tested its performance on $t$ only.
As expected, this setup appears challenging, causing severe performance drops, with accuracy on \textit{Count} and \textit{QueryAttribute} dropping to zero.
This stems from the non-overlapping labels spaces of \textit{Count} (digits 0 to 9) and \textit{QueryAttribute} (all attributes names) with the other groups (binary ``yes'' / ``no''). Thus, the model cannot predict these labels in a zero-shot transfer.



\subsection{Reasoning transfer in COG Canonical}
\label{sec:reasoning-transfer-cog}
\begin{figure}[b!]
	\centering
	\includegraphics[width=0.8\textwidth]{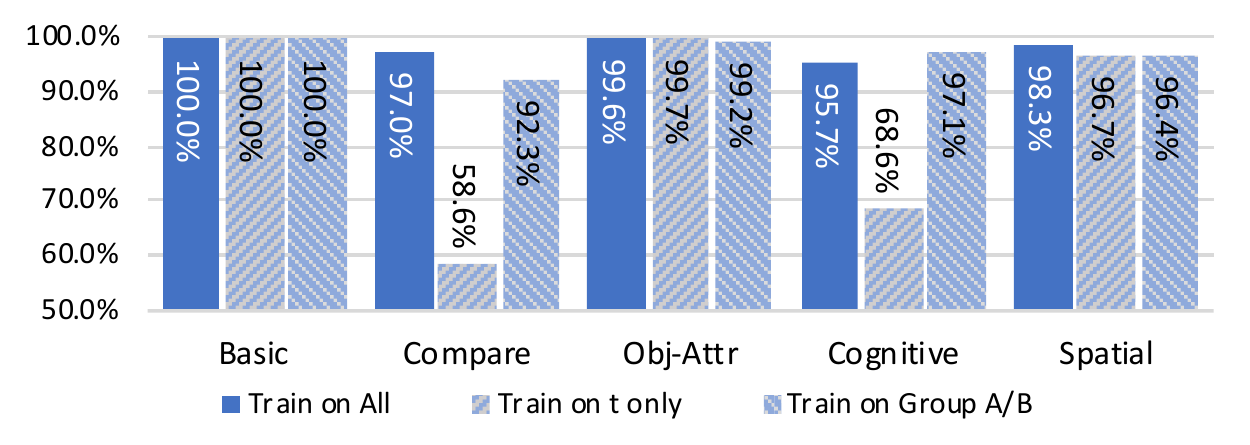}
	\caption{Accuracies of SAMNet when testing its reasoning transfer capabilities on the Canonical variant of COG.}
	\label{fig:cog_reasoning_transfer}
\end{figure}

In order to further investigate whether reasoning transfer is effective in leveraging information gained by training a task family at a higher level of the hierarchy, we also conducted a set of experiments using the Canonical variant of COG.
The order of the performed experiments is analogical to the one in the previous section.
In~\cref{fig:cog_reasoning_transfer}, the first bar in each cluster, ``Train on All'', is the baseline, i.e.
we train a single SAMNet instance on all tasks and test on each of the five task groups from the task hierarchy presented in \cref{fig:task-groups}.
Note that these results are weighted averages of the accuracies of our model on the Canonical variant tasks (\cref{fig:samnet_cog_detailed}).

Next, we experimented with the ``Train on t only'' setup, i.e. training and testing each of 5 SAMNet instances on a single task group $t$ (i.e. single leaf of the proposed hierarchy).
For \textit{Basic}, \textit{Spatial} and \textit{Obj-Attr}, the accuracy is matching the baseline (``Train on All''), suggesting that each group contain sufficient information for developing all necessary skills.
Therefore, models trained on these groups may not benefit from any additional training on other task groups.
However, the accuracy drops for \textit{Compary} and \textit{Cognitive}, suggesting the opposite.

Therefore, we have performed a final set of experiments (``Train on Group A/B''), where we trained two SAMNet instances: a) trained on all tasks from \textit{Group A} and tested on each task from the lowest groups (i.e. \textit{Basic}, \textit{Obj-Attr} and \textit{Compare}) separately; b) trained on all tasks from \textit{Group B} and tested separately on \textit{Spatial} and \textit{Cognitive}.
As expected, training on the other tasks from \textit{Group A} enabled the model to significantly improve on the \textit{Compare} task (from 58.6\% to 92.3\%).
A similar improvement (from 68.8\% to 97.1\%) was observed in the second experiment, when trained on \textit{Group B} and tested on \textit{Cognitive}.
Moreover, the achieved accuracy is in fact higher by 1.4 point than the one achieved when training on all tasks (95.7\%).
This suggests the model may have lacked capacity to master questions from all tasks and that training on a smaller subset allowed better performance on this subset.
Finally, we can interestingly note for \textit{Obj-Attr} and \textit{Spatial} that training on the group A/B does not appear to help. When comparing with ``Train on t only'', we observe slight accuracy drops (around 0.3-0.5 point).
Understanding this phenomenon warrants further investigations on the dependencies between the reasoning required by the particular questions and the transferred reasoning.


%% file: summary.tex
\section{Summary}


Even though transfer learning in computer vision is a common practice, a holistic view of its impact on visual reasoning was missing.
To capture and quantify the influence of transfer learning on visual reasoning, we proposed a new taxonomy, articulated around three aspects: feature, temporal and reasoning transfer.  This enabled us to reuse the existing datasets for image and video QA, namely CLEVR and COG, and isolate splits better capturing reasoning transfer.
We note that some of the proposed splits form tasks (e.g. Train on all but \textit{t}) which are complementary to well established ones in the literature, e.g. in Taskonomy~\cite{zamir2018taskonomy}.

Our experiments on transfer learning showed the shortcomings of existing approaches, especially for video reasoning.  Hence, we designed a novel Memory-Augmented Neural Network model called SAMNet, with mechanisms to address these deficiencies.
SAMNet showed significant improvements over state-of-the-art models on the COG dataset for Video Reasoning  and achieved comparable performance on the CLEVR dataset for Image Reasoning.
It also demonstrated excellent generalization capabilities for temporal and reasoning transfer. Moreover, through the cautious use of fine-tuning, SAMNet's performance advanced even further.
We hope that the proposed taxonomy, newly established reasoning transfer tasks, along with the provided baselines will bolster new research on both models and datasets for transfer learning on visual reasoning.

%% file: appendix.tex
\section{Datasets: CLEVR-CoGenT and COG}
\label{sec:datasets-desc}
%
The CoGenT dataset~\cite{johnson2017clevr} contains:
\begin{itemize}
	\def\labelitemi{--}
	\item Training set of 70,000 images and 699,960 questions in Condition A,
	\item Validation set of 15,000 images and 149,991 questions in Condition A,
	\item Test set of 15,000 images and 149,980 questions in Condition A (without answers),
	\item Validation set of 15,000 images and 150,000 questions in Condition B,
	\item Test set of 15,000 images and 149,992 questions in Condition B (without answers),
	\item Scene graphs and functional programs for all training/validation images/questions.
\end{itemize}

\noindent The combinations of attributes in CoGenT-A and CoGenT-B are shown in~\cref{tab:cogent_conditions_supplement}.
\begin{table}[ht]
	\centering
	\begin{tabular}{cccc}
		\toprule
		Dataset	&	Cubes	&	Cylinders	&	Spheres	\\
		\midrule
		CoGenT-A	&	Gray / Blue / Brown / Yellow	&	Red / Green / Purple / Cyan	&	Any color	\\
		CoGenT-B	&	Red / Green / Purple / Cyan	&	Gray / Blue / Brown / Yellow	&	Any color 	\\
		\bottomrule
	\end{tabular}
	\smallskip
	\caption{Colors \& shapes combinations present in CoGenT-A \& -B datasets.}
	\label{tab:cogent_conditions_supplement}
\end{table}

\noindent The Canonical and Hard variants of the COG dataset~\cite{yang2018dataset} are contrasted in~\cref{tab:cog_variants_supplement}.
\begin{table}[ht]
	\centering
	\begin{tabular}{lcccccc}
		\toprule
		Variant    &  	number &  	maximum & maximum & size & size & size  \\ 
		& of   & memory & number of & of & of & of  \\
		& frames & duration & distractors & training set & validation set & test set \\
		\midrule
		Canonical & 4 & 3 & 1 & 10000320 & 500016 & 500016 \\	
		Hard  & 8 & 7 & 10 & 10000320 & 500016  & 500016 \\
		\bottomrule	
	\end{tabular}
	\smallskip
	\caption{Details of the Canonical and Hard variants of the COG dataset.}
	\label{tab:cog_variants_supplement}
\end{table}
%

\clearpage
\section{Complete COG results}
\label{sec:cog-all-results}

\begin{table*}[htb]
	\centering
	\begin{adjustbox}{width=\textwidth}
		\begin{tabular}{l r r r r r r r r r r}
			\toprule[1.25pt]
			Model & \multicolumn{4}{c}{SAMNet} &  &\multicolumn{4}{c}{Baseline Model} \\
			\cmidrule(lr){2-5} 
			\cmidrule(lr){7-10} 
			&&&&& & paper & code & code & paper\\
			\cmidrule(lr){7-10} 
			Trained on       & canonical & canonical & canonical & hard &           &  canonical  & canonical  & canonical & hard \\ 
			Fine tuned on  & - & - & hard  & - &           & -   & - & hard & - \\ 
			Tested on        & canonical & hard & hard & hard &            &canonical  & hard & hard & hard  \\ 
			\midrule[1pt]	
			Overall accuracy & 98.0 & 91.6 & 96.5  & 96.1 &           & 97.6  & 65.9 & 78.1& 80.1 \\ 
			\midrule[1pt]	
			AndCompareColor			&	93.5	&	82.7	&	89.2	&	80.6	&	&	81.9	&	57.1	&	60.7	&	51.4	 \\
			AndCompareShape			&	93.2	&	83.7	&	89.7	&	80.1	&	&	80.0	&	53.1	&	50.3	&	50.7 \\
			AndSimpleCompareColor		&	99.2	&	85.3	&	97.6	&	99.4	&	&	99.7	&	53.4	&	77.1	&	78.2 \\
			AndSimpleCompareShape		&	99.2	&	85.8	&	97.6	&	99.2	&	&	100.0	&	56.7	&	79.3	&	77.9 \\
			CompareColor			&	98.1	&	89.3	&	95.9	&	99.7	&	&	99.2	&	56.1	&	67.9	&	50.1 \\
			CompareShape			&	98.0	&	89.7	&	95.9	&	99.2	&	&	99.4	&	66.8	&	65.4	&	50.5	 \\
			Exist				&	100.0	&	99.7	&	99.8	&	99.8	&	&	100.0	&	63.5	&	96.1	&	99.3 \\
			ExistColor			&	100.0	&	99.6	&	99.9	&	99.9	&	&	99.0	&	70.9	&	99	&	89.8 \\
			ExistColorOf			&	99.9	&	95.5	&	99.7	&	99.8	&	&	99.7	&	51.5	&	76.1	&	73.1 \\
			ExistColorSpace			&	94.1	&	88.8	&	91.0	&	90.8	&	&	98.9	&	72.8	&	77.3	&	89.2 \\
			ExistLastColorSameShape		&	99.5	&	99.4	&	99.4	&	98.0	&	&	100.0	&	65.0	&	62.5	&	50.4 \\
			ExistLastObjectSameObject	&	97.3	&	97.5	&	97.7	&	97.5	&	&	98.0	&	77.5	&	61.7	&	60.2 \\
			ExistLastShapeSameColor		&	98.2	&	98.5	&	98.8	&	97.5	&	&	100.0	&	87.8	&	60.4	&	50.3 \\
			ExistShape			&	100.0	&	99.5	&	100.0	&	100.0	&	&	100.0	&	77.1	&	98.2	&	92.5 \\
			ExistShapeOf			&	99.4	&	95.9	&	99.2	&	99.2	&	&	100.0	&	52.7	&	74.7	&	72.70 \\
			ExistShapeSpace			&	93.4	&	87.5	&	91.1	&	90.5	&	&	97.7	&	70.0	&	89.8	&	89.80 \\
			ExistSpace			&	95.3	&	89.7	&	93.2	&	93.3	&	&	98.9	&	71.1	&	88.1	&	92.8 \\
			GetColor			&	100.0	&	95.8	&	99.9	&	100.0	&	&	100.0	&	71.4	&	83.1	&	97.9 \\
			GetColorSpace			&	98.0	&	90.0	&	95.0	&	95.4	&	&	98.2	&	71.8	&	73.	&	92.3 \\
			GetShape			&	100.0	&	97.3	&	99.9	&	99.9	&	&	100.0	&	83.5	&	89.2	&	97.1	 \\
			GetShapeSpace			&	97.5	&	89.4	&	93.9	&	94.3	&	&	98.1	&	78.7	&	77.3	&	90.3 \\
			SimpleCompareShape		&	99.9	&	91.4	&	99.7	&	99.9	&	&	100.0	&	67.7	&	96.7	&	99.3 \\
			SimpleCompareColor		&	100.0	&	91.6	&	99.80	&	99.9	&	&	100.0	&	64.2	&	90.4	&	99.3 \\
			\bottomrule[1.25pt]
		\end{tabular}
	\end{adjustbox}
	\smallskip
	\caption{COG test set accuracies for SAMNet and the baseline model~\cite{yang2018dataset}. For the baseline model,
		`paper' denotes results reproduced from~\cite{yang2018dataset} along with some further clarifications by the authors (private communication)  regarding the performance on individual task types while `code' denotes results of our experiments using their implementation~\cite{yang2018implement}.}
	\label{tab:all-results}
\end{table*}

\clearpage
\section{Complete CLEVR-CoGenT results}
\label{sec:full-cogent-results}

\begin{table*}[htb]
	\centering
	\begin{adjustbox}{width=\textwidth}
	\begin{tabular}{c c c c c c c }\toprule
		\multicolumn{1}{c }{\textbf{}} & \multicolumn{6}{c }{\textbf{Test Accuracy (\%)}} \\  
		\multicolumn{1}{c }{\textbf{}} & \multicolumn{6}{c }{on \texttt{valA} -- 150,000 samples} \\ 
		\cmidrule(lr){2-7}
		\multicolumn{1}{ c }{\textbf{Experiments}} & \textbf{Overall}& \textbf{Exist}  & \textbf{Count} & \textbf{CompareInteger} & \textbf{CompareAttribute} & \textbf{QueryAttribute}\\ 
		\cmidrule(lr){1-1}
		\cmidrule(lr){2-7}
		\multicolumn{1}{ c }{Exist only} & 26.07 & \emph{74.20}	& 0.0	& 59.79	& 59.15 & 0.0 \\ 
		\multicolumn{1}{ c }{Count only} & 14.89  & 0.0	& \emph{62.78}	& 0.0 & 0.0 & 0.0 \\ 
		\multicolumn{1}{ c }{CompareInteger only} & 23.44 & 48.15	& 0.0	& \emph{77.98}	& 54.08 & 0.0 \\ 
		\multicolumn{1}{ c }{CompareAttribute only} & 23.50 & 51.93	& 0.0 & 59.06 & \emph{61.73} & 0.0 \\ 
		\multicolumn{1}{ c }{QueryAttribute only} & 34.64 	& 0.0	& 0.0	& 0.0 & 0.0 & \emph{97.08} \\ 		
		\cmidrule(lr){1-1}
		\cmidrule(lr){2-7}
		
		\multicolumn{1}{ c }{All tasks} & 95.32 & 98.4 	& 86.75	& 96.0	& 98.65	& 98.02 \\ 
		\cmidrule(lr){1-1}
		\cmidrule(lr){2-7}
		
		\multicolumn{1}{ c }{All tasks but Exist} & 90.33 	& \emph{60.42}	& 86.12	& 96.18	& 98.52 & 98.60 \\ 
		\multicolumn{1}{ c }{All tasks but Count} & 74.59 	& 97.51	& \emph{0.0}	& 94.37	& 98.42 & 98.47 \\ 
		\multicolumn{1}{ c }{All tasks but CompareInteger} & 91.53 	& 98.22	& 86.09	& \emph{56.35}	& 98.78 & 98.40 \\ 
		\multicolumn{1}{ c }{All tasks but CompareAttribute} & 81.92 	& 98.32	& 86.36	& 95.86	& \emph{23.18} & 98.44 \\ 
		\multicolumn{1}{ c }{All tasks but QueryAttribute} & 42.17 	& 76.79	& 54.64	& 79.87 & 64.13 & \emph{0.0} \\ 
		
		\cmidrule(lr){1-1}
		\cmidrule(lr){2-7}
		
		\multicolumn{1}{ c }{\textit{Trained on all tasks -- Finetune on $t$}} & &  & &  & &\\ 
		\multicolumn{1}{ c }{Exist} & 94.16 & \emph{98.04} & 82.96 & 95.02 & 98.12 & 97.97 \\ 
		\multicolumn{1}{ c }{Count} & 95.10 & 96.46 & \emph{88.28} & 94.78 & 97.01 & 98.26 \\
		\multicolumn{1}{ c }{CompareInteger} & 95.31 & 98.39 & 86.56 & \emph{96.07} & 98.70 & 98.09 \\
		\multicolumn{1}{ c }{CompareAttribute} & 95.31 & 98.40 & 86.78 & 96.00 & \emph{98.66} & 98.04 \\
		\multicolumn{1}{ c }{QueryAttribute} & 93.49 & 97.45 & 82.09 & 92.23 & 97.85& \emph{97.76} \\
		
		\bottomrule
	\end{tabular}
	\end{adjustbox}
    \smallskip
	\caption{Complete set of results for SAMNet on CLEVR-CoGenT.}
	\label{tab:CoGenT-results}
\end{table*}

%% file: ms.bbl
\newcommand{\etalchar}[1]{$^{#1}$}
\begin{thebibliography}{JHvdM{\etalchar{+}}17b}

\bibitem[BCB14]{bahdanau2014neural}
Dzmitry Bahdanau, Kyunghyun Cho, and Yoshua Bengio.
\newblock Neural machine translation by jointly learning to align and
  translate.
\newblock {\em arXiv preprint arXiv:1409.0473}, 2014.

\bibitem[DCLT18]{devlin2018bert}
Jacob Devlin, Ming-Wei Chang, Kenton Lee, and Kristina Toutanova.
\newblock Bert: Pre-training of deep bidirectional transformers for language
  understanding.
\newblock {\em arXiv preprint arXiv:1810.04805}, 2018.

\bibitem[DDS{\etalchar{+}}09]{deng2009imagenet}
Jia Deng, Wei Dong, Richard Socher, Li-Jia Li, Kai Li, and Li~Fei-Fei.
\newblock Imagenet: A large-scale hierarchical image database.
\newblock In {\em 2009 IEEE conference on computer vision and pattern
  recognition}, pages 248--255. Ieee, 2009.

\bibitem[Gan18]{yang2018implement}
Igor Ganichev.
\newblock Cog implementation.
\newblock \url{https://github.com/google/cog}, 2018.

\bibitem[GMH13]{graves2013speech}
Alex Graves, Abdel-rahman Mohamed, and Geoffrey Hinton.
\newblock Speech recognition with deep recurrent neural networks.
\newblock In {\em 2013 IEEE international conference on acoustics, speech and
  signal processing}, pages 6645--6649. IEEE, 2013.

\bibitem[GWD14]{graves2014neural}
Alex Graves, Greg Wayne, and Ivo Danihelka.
\newblock Neural turing machines.
\newblock {\em arXiv preprint arXiv:1410.5401}, 2014.

\bibitem[GWR{\etalchar{+}}16]{graves2016hybrid}
Alex Graves, Greg Wayne, Malcolm Reynolds, Tim Harley, Ivo Danihelka, Agnieszka
  Grabska-Barwi{\'n}ska, Sergio~G{\'o}mez Colmenarejo, Edward Grefenstette,
  Tiago Ramalho, John Agapiou, et~al.
\newblock Hybrid computing using a neural network with dynamic external memory.
\newblock {\em Nature}, 538(7626):471, 2016.

\bibitem[HAE16]{huh2016makes}
Minyoung Huh, Pulkit Agrawal, and Alexei~A Efros.
\newblock What makes imagenet good for transfer learning?
\newblock {\em arXiv preprint arXiv:1608.08614}, 2016.

\bibitem[HM18]{hudson2018compositional}
Drew~A. Hudson and Christopher~D. Manning.
\newblock Compositional attention networks for machine reasoning.
\newblock {\em International Conference on Learning Representations}, 2018.

\bibitem[HRS19]{haurilet2019s}
Monica Haurilet, Alina Roitberg, and Rainer Stiefelhagen.
\newblock It's not about the journey; it's about the destination: Following
  soft paths under question-guidance for visual reasoning.
\newblock In {\em Proceedings of the IEEE Conference on Computer Vision and
  Pattern Recognition}, pages 1930--1939, 2019.

\bibitem[HS97]{hochreiter1997long}
Sepp Hochreiter and J{\"u}rgen Schmidhuber.
\newblock Long short-term memory.
\newblock {\em Neural computation}, 9(8):1735--1780, 1997.

\bibitem[HZRS16]{he2016deep}
Kaiming He, Xiangyu Zhang, Shaoqing Ren, and Jian Sun.
\newblock Deep residual learning for image recognition.
\newblock In {\em Proceedings of the IEEE conference on computer vision and
  pattern recognition}, pages 770--778, 2016.

\bibitem[IS18]{iglovikov2018ternausnet}
Vladimir Iglovikov and Alexey Shvets.
\newblock Ternausnet: U-net with vgg11 encoder pre-trained on imagenet for
  image segmentation.
\newblock {\em arXiv preprint arXiv:1801.05746}, 2018.

\bibitem[JBM{\etalchar{+}}18]{jayram2018learning}
TS~Jayram, Younes Bouhadjar, Ryan~L McAvoy, Tomasz Kornuta, Alexis Asseman,
  Kamil Rocki, and Ahmet~S Ozcan.
\newblock Learning to remember, forget and ignore using attention control in
  memory.
\newblock {\em arXiv preprint arXiv:1809.11087}, 2018.

\bibitem[JHvdM{\etalchar{+}}17a]{johnson2017clevr}
Justin Johnson, Bharath Hariharan, Laurens van~der Maaten, Li~Fei-Fei,
  C~Lawrence~Zitnick, and Ross Girshick.
\newblock Clevr: A diagnostic dataset for compositional language and elementary
  visual reasoning.
\newblock In {\em Proceedings of the IEEE Conference on Computer Vision and
  Pattern Recognition}, pages 2901--2910, 2017.

\bibitem[JHvdM{\etalchar{+}}17b]{johnson2017inferring}
Justin Johnson, Bharath Hariharan, Laurens van~der Maaten, Judy Hoffman,
  Li~Fei-Fei, C~Lawrence~Zitnick, and Ross Girshick.
\newblock Inferring and executing programs for visual reasoning.
\newblock In {\em Proceedings of the IEEE International Conference on Computer
  Vision}, pages 2989--2998, 2017.

\bibitem[KMM{\etalchar{+}}18]{kornuta2018accelerating}
Tomasz Kornuta, Vincent Marois, Ryan~L McAvoy, Younes Bouhadjar, Alexis
  Asseman, Vincent Albouy, TS~Jayram, and Ahmet~S Ozcan.
\newblock Accelerating machine learning research with mi-prometheus.
\newblock In {\em NeurIPS Workshop on Machine Learning Open Source Software
  (MLOSS)}, volume 2018, 2018.

\bibitem[KRS{\etalchar{+}}19]{kornuta2019leveraging}
Tomasz Kornuta, Deepta Rajan, Chaitanya Shivade, Alexis Asseman, and Ahmet~S
  Ozcan.
\newblock Leveraging medical visual question answering with supporting facts.
\newblock {\em arXiv preprint arXiv:1905.12008}, 2019.

\bibitem[KSH12]{krizhevsky2012imagenet}
Alex Krizhevsky, Ilya Sutskever, and Geoffrey~E Hinton.
\newblock Imagenet classification with deep convolutional neural networks.
\newblock In {\em Advances in neural information processing systems}, pages
  1097--1105, 2012.

\bibitem[KSL19]{kornblith2019better}
Simon Kornblith, Jonathon Shlens, and Quoc~V Le.
\newblock Do better imagenet models transfer better?
\newblock In {\em Proceedings of the IEEE Conference on Computer Vision and
  Pattern Recognition}, pages 2661--2671, 2019.

\bibitem[LBH15]{lecun2015deep}
Yann LeCun, Yoshua Bengio, and Geoffrey Hinton.
\newblock {Deep Learning}.
\newblock {\em Nature}, 521(7553):436, 2015.

\bibitem[MJA{\etalchar{+}}18]{marois2018transfer}
Vincent Marois, TS~Jayram, Vincent Albouy, Tomasz Kornuta, Younes Bouhadjar,
  and Ahmet~S Ozcan.
\newblock On transfer learning using a {MAC} model variant.
\newblock In {\em NeurIPS'18 Visually-Grounded Interaction and Language (ViGIL)
  Workshop}, 2018.

\bibitem[MKK19]{mogadala2019trends}
Aditya Mogadala, Marimuthu Kalimuthu, and Dietrich Klakow.
\newblock Trends in integration of vision and language research: A survey of
  tasks, datasets, and methods.
\newblock {\em arXiv preprint arXiv:1907.09358}, 2019.

\bibitem[MSC{\etalchar{+}}13]{mikolov2013distributed}
Tomas Mikolov, Ilya Sutskever, Kai Chen, Greg~S Corrado, and Jeff Dean.
\newblock Distributed representations of words and phrases and their
  compositionality.
\newblock In {\em Advances in neural information processing systems}, pages
  3111--3119, 2013.

\bibitem[MTSM18]{mascharka2018transparency}
David Mascharka, Philip Tran, Ryan Soklaski, and Arjun Majumdar.
\newblock Transparency by design: Closing the gap between performance and
  interpretability in visual reasoning.
\newblock In {\em Proceedings of the IEEE conference on computer vision and
  pattern recognition}, pages 4942--4950, 2018.

\bibitem[PGC{\etalchar{+}}17]{paszke2017automatic}
Adam Paszke, Sam Gross, Soumith Chintala, Gregory Chanan, Edward Yang, Zachary
  DeVito, Zeming Lin, Alban Desmaison, Luca Antiga, and Adam Lerer.
\newblock Automatic differentiation in {PyTorch}.
\newblock {\em NIPS 2017 Workshop Autodiff}, 2017.

\bibitem[PNI{\etalchar{+}}18]{peters2018deep}
Matthew~E Peters, Mark Neumann, Mohit Iyyer, Matt Gardner, Christopher Clark,
  Kenton Lee, and Luke Zettlemoyer.
\newblock Deep contextualized word representations.
\newblock {\em arXiv preprint arXiv:1802.05365}, 2018.

\bibitem[PSDV{\etalchar{+}}18]{perez2018film}
Ethan Perez, Florian Strub, Harm De~Vries, Vincent Dumoulin, and Aaron
  Courville.
\newblock Film: Visual reasoning with a general conditioning layer.
\newblock In {\em Thirty-Second AAAI Conference on Artificial Intelligence},
  2018.

\bibitem[PSM14]{pennington2014glove}
Jeffrey Pennington, Richard Socher, and Christopher Manning.
\newblock Glove: Global vectors for word representation.
\newblock In {\em Proceedings of the 2014 conference on empirical methods in
  natural language processing (EMNLP)}, pages 1532--1543, 2014.

\bibitem[PY09]{pan2009survey}
Sinno~Jialin Pan and Qiang Yang.
\newblock A survey on transfer learning.
\newblock {\em IEEE Transactions on knowledge and data engineering},
  22(10):1345--1359, 2009.

\bibitem[RDGF16]{redmon2016you}
Joseph Redmon, Santosh Divvala, Ross Girshick, and Ali Farhadi.
\newblock You only look once: Unified, real-time object detection.
\newblock In {\em Proceedings of the IEEE conference on computer vision and
  pattern recognition}, pages 779--788, 2016.

\bibitem[Rip93]{ripley1993statistical}
Brian~D Ripley.
\newblock Statistical aspects of neural networks.
\newblock {\em Networks and chaos—statistical and probabilistic aspects},
  50:40--123, 1993.

\bibitem[SSCH18]{song2018explore}
Xiaomeng Song, Yucheng Shi, Xin Chen, and Yahong Han.
\newblock Explore multi-step reasoning in video question answering.
\newblock In {\em 2018 ACM Multimedia Conference on Multimedia Conference},
  pages 239--247. ACM, 2018.

\bibitem[SZ14]{simonyan2014very}
Karen Simonyan and Andrew Zisserman.
\newblock Very deep convolutional networks for large-scale image recognition.
\newblock {\em arXiv preprint arXiv:1409.1556}, 2014.

\bibitem[VSP{\etalchar{+}}17]{vaswani2017attention}
Ashish Vaswani, Noam Shazeer, Niki Parmar, Jakob Uszkoreit, Llion Jones,
  Aidan~N Gomez, {\L}ukasz Kaiser, and Illia Polosukhin.
\newblock Attention is all you need.
\newblock In {\em Advances in Neural Information Processing Systems}, pages
  5998--6008, 2017.

\bibitem[WCB14]{weston2014memory}
Jason Weston, Sumit Chopra, and Antoine Bordes.
\newblock Memory networks.
\newblock {\em arXiv preprint arXiv:1410.3916}, 2014.

\bibitem[WKW16]{weiss2016survey}
Karl Weiss, Taghi~M Khoshgoftaar, and DingDing Wang.
\newblock A survey of transfer learning.
\newblock {\em Journal of Big data}, 3(1):9, 2016.

\bibitem[WM96]{warner1996understanding}
Brad Warner and Manavendra Misra.
\newblock Understanding neural networks as statistical tools.
\newblock {\em The american statistician}, 50(4):284--293, 1996.

\bibitem[YGW{\etalchar{+}}18]{yang2018dataset}
Guangyu~Robert Yang, Igor Ganichev, Xiao-Jing Wang, Jonathon Shlens, and David
  Sussillo.
\newblock A dataset and architecture for visual reasoning with a working
  memory.
\newblock In {\em European Conference on Computer Vision}, pages 729--745.
  Springer, 2018.

\bibitem[ZSS{\etalchar{+}}18]{zamir2018taskonomy}
Amir~R Zamir, Alexander Sax, William Shen, Leonidas~J Guibas, Jitendra Malik,
  and Silvio Savarese.
\newblock Taskonomy: Disentangling task transfer learning.
\newblock In {\em Proceedings of the IEEE Conference on Computer Vision and
  Pattern Recognition}, pages 3712--3722, 2018.

\end{thebibliography}
